\documentclass{article}
\usepackage{spconf,amsmath,graphicx,hyperref,tabularx,booktabs,arydshln}
\usepackage{comment}
\usepackage{subcaption}

\title{Efficient Segment Anything with Depth-Aware Fusion and Limited Training Data}
\name{
Yiming Zhou$^{123}$\thanks{This work was supported by htw saar through the EXIST-Forschungstransfer program (Project No. 03EFSL0034).},
Xuanjie Xie$^{2}$,
Panfeng Li$^{4}$,
Albrecht Kunz$^{2}$,
Ahmad Osman$^{23}$,
Xavier Maldague$^{1}$
}
\address{
$^{1}$ Université Laval, Quebec City, Canada \\
$^{2}$ Saarland University of Applied Sciences, Saarbruecken, Germany \\
$^{3}$ Fraunhofer Institute for Nondestructive Testing IZFP, Saarbruecken, Germany \\
$^{4}$ University of Michigan, Ann Arbor, USA 
}

\begin{document}
%\ninept
%
\maketitle
\begin{abstract}
Segment Anything Models (SAM) achieve impressive universal segmentation performance but require massive datasets (e.g., 11M images) and rely solely on RGB inputs. Recent efficient variants reduce computation but still depend on large-scale training. We propose a lightweight RGB-D fusion framework that augments EfficientViT-SAM with monocular depth priors. Depth maps are generated with a pretrained estimator and fused mid-level with RGB features through a dedicated depth encoder. Trained on only 11.2k samples (less than 0.1\% of SA-1B), our method achieves higher accuracy than EfficientViT-SAM, showing that depth cues provide strong geometric priors for segmentation. 
Our results demonstrate that depth cues enable data-efficient and lightweight segmentation suitable for resource-limited scenarios.
\end{abstract}
\section{Introduction}
\label{sec:intro}
The Segment Anything Model (SAM) \cite{kirillov2023segment}  has recently demonstrated strong zero-shot generalization in tasks such as AR/VR, medical imaging, and interactive editing. Its success is largely attributed to large-scale pretraining on SA-1B \cite{kirillov2023segment} dataset which contains 11M images and 1B masks. However, SAM’s reliance on massive datasets and heavy transformer-based encoders makes it computationally expensive and unsuitable for real-time or resource-constrained applications.

To mitigate these limitations, lightweight variants \cite{zhang2023faster, zhao2023fast, liu2023efficientvit} have been proposed. These models reduce inference cost by simplifying the image encoder, but remain restricted to RGB inputs and still require large-scale training data to reach competitive accuracy. This reliance on massive annotated datasets limits their scalability in domains with scarce data.

In this work, we propose \textbf{Depth-Aware EfficientViT-SAM}, a lightweight segmentation framework that augments EfficientViT-SAM \cite{zhang2024efficientvit} with geometric priors from monocular depth estimation. Specifically, we generate depth maps using a pretrained estimator and introduce a dedicated depth encoder that is fused mid-level with the RGB encoder. Depth cues provide strong geometric priors, particularly for refining object boundaries, enabling data-efficient training without retraining a large transformer backbone.

Trained on only 11.2k images ($<$0.1\% of SA-1B), our framework surpasses EfficientViT-SAM on zero-shot benchmarks, demonstrating substantial gains in accuracy despite the tiny training set. These results highlight the effectiveness of incorporating geometric depth priors for both data-efficient training and lightweight inference.

In summary, our contributions are as follows:
\begin{itemize}
    \item We propose \textbf{Depth-Aware EfficientViT-SAM}, which integrates monocular depth cues into EfficientViT-SAM to enhance boundary segmentation.
    \item Our model is trained on only 11.2k SA-1B images ($<$0.1\% of the dataset) for 4 epochs, yet achieves strong zero-shot performance with lightweight training.  
    \item We design a loss function combining mask, dice, and IoU terms with auxiliary direct prediction losses, further enhancing segmentation accuracy.  
\end{itemize}

\begin{figure*}[t] % 'h' means here
    \centering
    \includegraphics[width=0.90\textwidth,height=0.28\textheight]{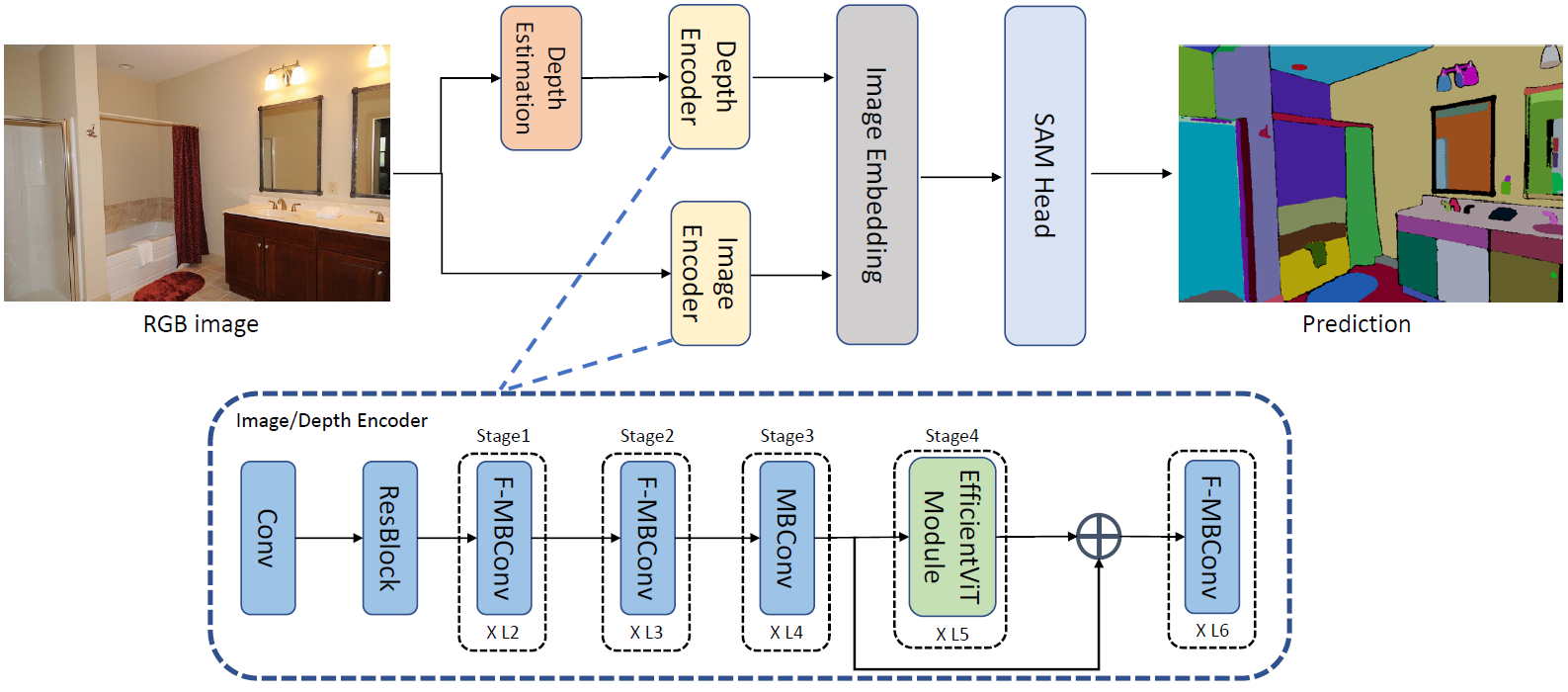}
    \caption{Overview of our framework. Depth maps are estimated with DepthAnything \cite{yang2024depth} and encoded alongside RGB features using identical encoder architectures. Their fused embedding is processed by the SAM head, consisting of a prompt encoder and mask decoder, to produce the final segmentation.}
    \label{fig:pipeline}
\end{figure*}

\section{Related Work}
\label{sec:related work}

\subsection{Efficient Segment Anything Models}
The Segment Anything Model (SAM) \cite{kirillov2023segment} is a major advance in universal segmentation, achieving strong zero-shot performance with minimal user input. It has been applied to diverse tasks such as object tracking \cite{cheng2023segment}, medical imaging, and accessibility, and more recently extended to 3D perception tasks like reconstruction \cite{wang2025vggt, 10704527} and SLAM \cite{10904342, 11127324}, where depth cues are crucial for reliable scene understanding. However, SAM’s heavy transformer backbone and massive pretraining make it computationally expensive and difficult to deploy in resource-limited scenarios.

Several efficient variants address this issue: FastSAM \cite{zhao2023fast} replaces the transformer with a convolutional encoder, while MobileSAM \cite{zhang2023faster} applies distillation for edge devices. EfficientViT \cite{liu2023efficientvit} further accelerates SAM with a lightweight vision transformer. Building on this line of work, we adopt EfficientViT as the backbone and integrate depth maps to guide boundary refinement, aiming for both efficiency and improved accuracy.

\subsection{RGB-D Segmentation}
RGB-only segmentation often fails in textureless regions or near ambiguous boundaries. Depth provides complementary geometric cues that improve edge delineation and structural consistency. Recent RGB-D approaches \cite{jia2024geminifusion, yin2023dformer, yin2025dformerv2} exploit this by fusing depth within encoder–decoder or transformer frameworks. For example, DFormer \cite{yin2023dformer} introduced a unified RGB–depth backbone to reduce dual-stream overhead, while DFormerV2 \cite{yin2025dformerv2} leveraged explicit geometric priors for stronger reasoning. Advances in monocular depth estimation, such as DepthAnything \cite{yang2024depth}, now make high-quality depth maps readily accessible, enabling practical RGB-D integration in segmentation pipelines. These developments motivate our design of a depth-aware EfficientViT-SAM that explicitly fuses depth with RGB features to enhance boundary refinement and data efficiency.

\begin{figure*}[t] % use figure* if you want across 2 columns
    \centering
    % First row
    \begin{subfigure}[b]{0.24\textwidth}
        \includegraphics[width=\textwidth]{images/inputs/indoor1.pdf}
        \label{fig:input}
    \end{subfigure}\vspace{-0.3em}
    \begin{subfigure}[b]{0.24\textwidth}
        \includegraphics[width=\textwidth]{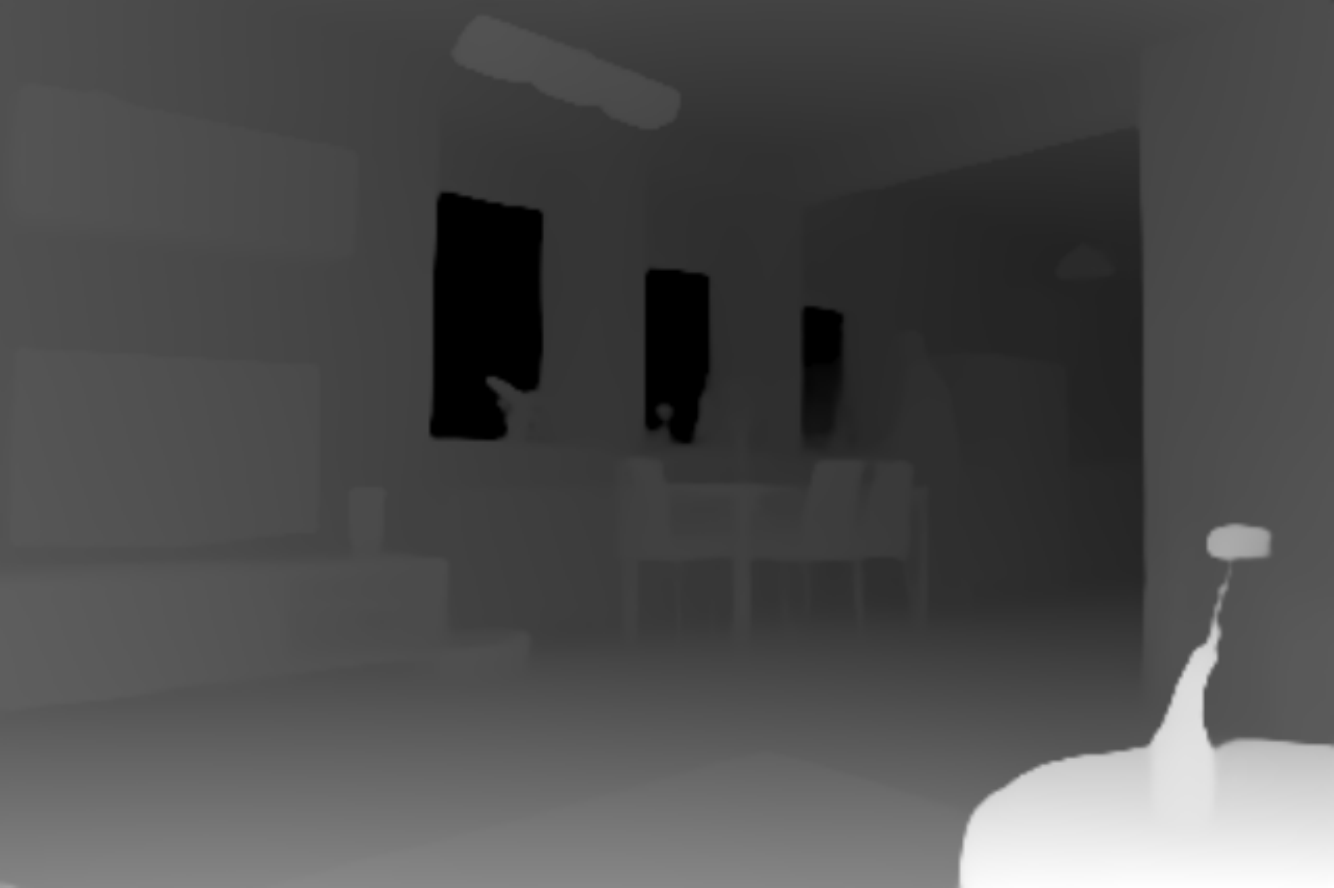}
        \label{fig:depth}
    \end{subfigure}\vspace{-0.3em}
    % Second row
    \begin{subfigure}[b]{0.24\textwidth}
        \includegraphics[width=\textwidth]{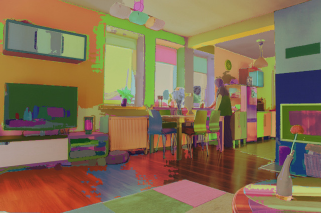}
        \label{fig:segA}
    \end{subfigure}\vspace{-0.3em}
    \begin{subfigure}[b]{0.24\textwidth}
        \includegraphics[width=\textwidth]{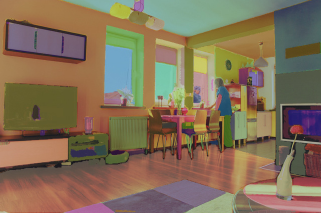}
        \label{fig:segB}
    \end{subfigure}\vspace{-0.3em}

    % ---------- Row 2 ----------
    \begin{subfigure}[b]{0.24\textwidth}
        \includegraphics[width=\textwidth]{images/inputs/indoor2.pdf}
        \label{fig:input_row2}
    \end{subfigure}\vspace{-0.3em}
    \begin{subfigure}[b]{0.24\textwidth}
        \includegraphics[width=\textwidth]{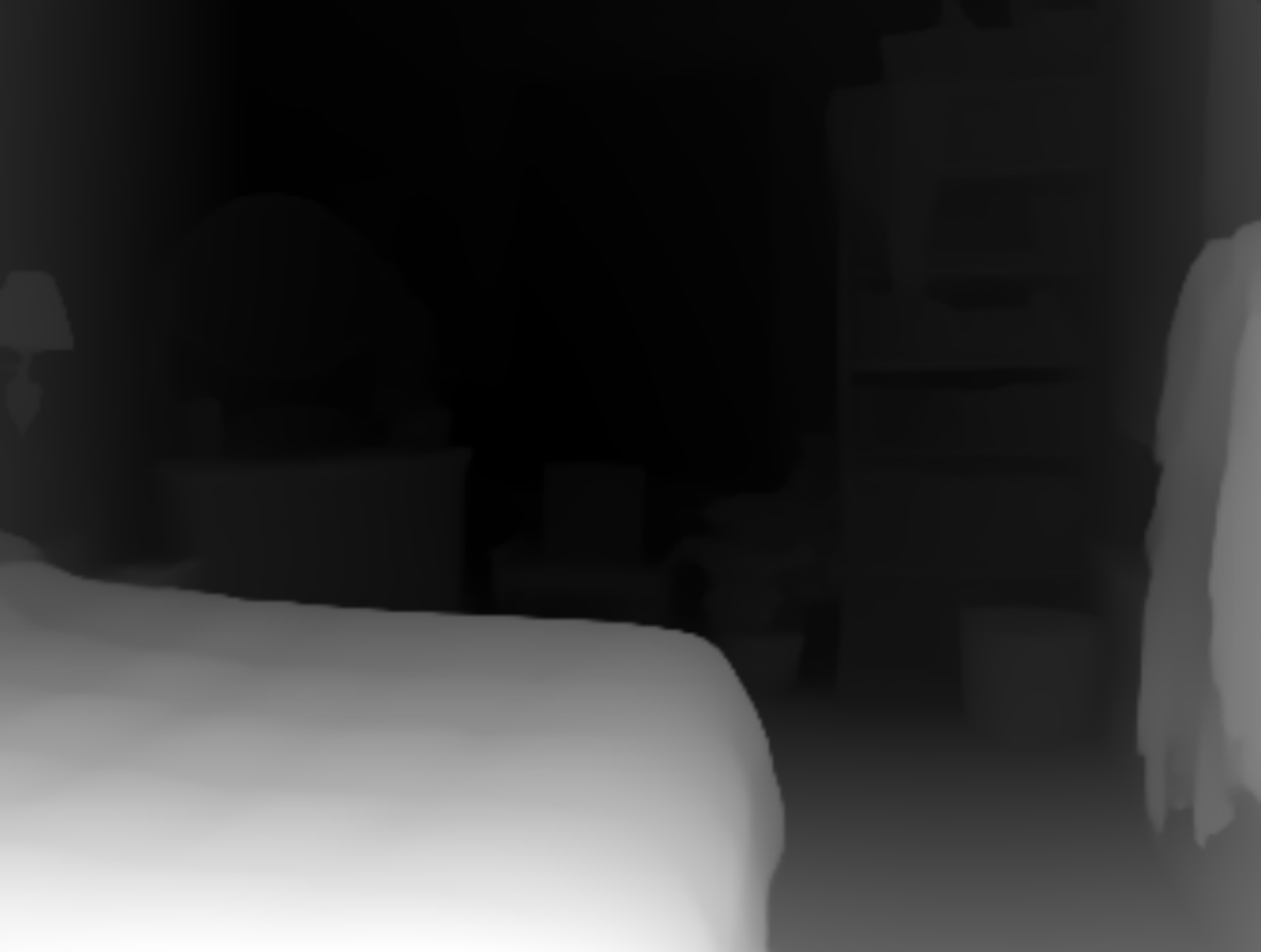}
        \label{fig:depth_row2}
    \end{subfigure}\vspace{-0.3em}
    \begin{subfigure}[b]{0.24\textwidth}
        \includegraphics[width=\textwidth]{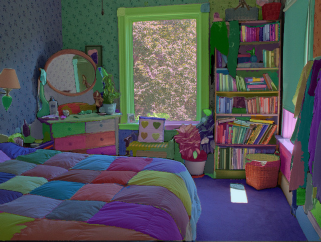}
        \label{fig:segA_row2}
    \end{subfigure}\vspace{-0.3em}
    \begin{subfigure}[b]{0.24\textwidth}
        \includegraphics[width=\textwidth]{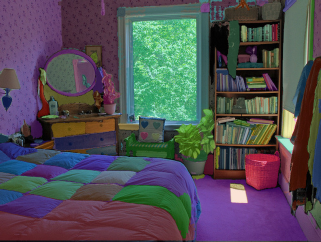}
        \label{fig:segB_row2}
    \end{subfigure}\vspace{-0.3em}

        % ---------- Row 3 ----------
    \begin{subfigure}[b]{0.24\textwidth}
        \includegraphics[width=\textwidth]{images/inputs/indoor4.pdf}
        \caption{Input Image}
        \label{fig:input_row3}
    \end{subfigure}
    \begin{subfigure}[b]{0.24\textwidth}
        \includegraphics[width=\textwidth]{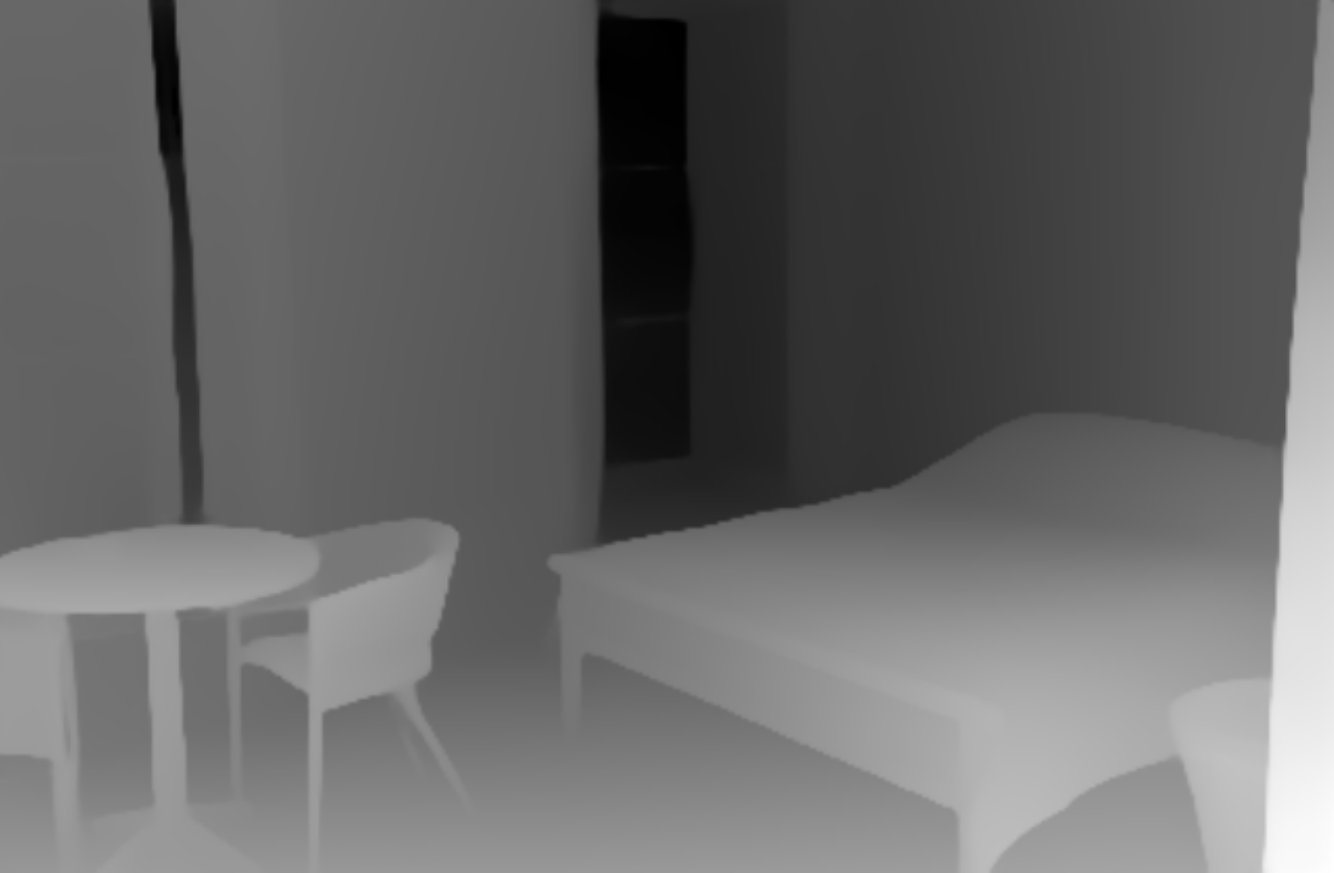}
        \caption{Depth}
        \label{fig:depth_row3}
    \end{subfigure}
    \begin{subfigure}[b]{0.24\textwidth}
        \includegraphics[width=\textwidth]{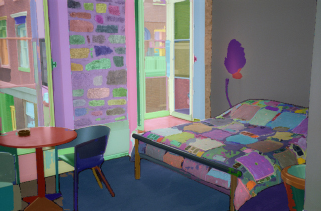}
        \caption{EfficientViT-SAM}
        \label{fig:segA_row3}
    \end{subfigure}
    \begin{subfigure}[b]{0.24\textwidth}
        \includegraphics[width=\textwidth]{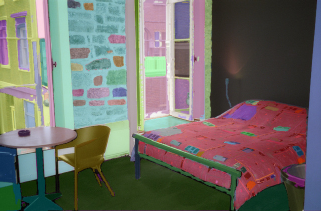}
        \caption{Our method}
        \label{fig:segB_row3}
    \end{subfigure}

    \caption{Qualitative comparison of point-prompted segmentation results. Each row shows one example, with (a) the input image, (b) the estimated depth map, (c) the segmentation produced by EfficientViT-SAM, and (d) the segmentation produced by our Depth-Aware EfficientViT-SAM.}

    \label{fig:comparison}
\end{figure*}

\section{Method}
\label{sec:method}
We propose \textbf{Depth-Aware EfficientViT-SAM}, which augments EfficientViT-SAM with a depth-aware encoder and a simple fusion mechanism. We keep the original SAM prompt encoder and mask decoder unchanged, while replacing SAM's heavy image encoder with EfficientViT to achieve efficiency.

\subsection{Model Architecture}
Figure~\ref{fig:pipeline} illustrates the framework. The RGB image encoder follows EfficientViT, consisting of four stages: early convolutional and residual blocks extract low-level features, intermediate stages employ F-MBConv and MBConv layers, and the final stage uses EfficientViT modules for high-level representations. 

In parallel, a depth encoder with the same architecture processes depth maps generated by DepthAnything~\cite{yang2024depth}. Depth images are replicated across three channels to match the RGB input dimension. Both encoders have separate parameters.  

After passing through the encoders, the resulting feature maps are fused before entering the SAM head. Let $F^{\mathrm{rgb}}$ and $F^{\mathrm{dep}}$ denote the image and depth embeddings. We adopt a direct additive fusion with a learnable scaling parameter $\alpha$:  
\begin{equation}
F^{\mathrm{fuse}} = F^{\mathrm{rgb}} + \alpha \cdot F^{\mathrm{dep}}
\end{equation}
This design preserves the dominant RGB features while allowing the network to adaptively leverage depth cues. The fused embedding is then processed by SAM’s original prompt encoder and mask decoder, enabling interactive segmentation with depth-aware representations.

\subsection{Loss Function}
We supervise segmentation with a weighted combination of binary cross-entropy and Dice loss:
\begin{equation}
\mathcal{L}_{\text{orig}} = \lambda_{\text{mask}} \mathcal{L}_{\text{mask}} + \lambda_{\text{dice}} \mathcal{L}_{\text{dice}},
\end{equation}
where $\lambda_{\text{mask}}{=}20$ and $\lambda_{\text{dice}}{=}1$.  
This baseline encourages pixel-wise accuracy while also improving overlap between predicted and ground-truth masks.

To further enhance boundary precision and prediction reliability, we extend the objective with three auxiliary terms: an IoU regression loss $\mathcal{L}_{\text{iou}}$, which guides the model to estimate mask quality; a direct supervision loss $\mathcal{L}_{\text{direct}}$ applied to intermediate predictions to stabilize training; and a boundary-focused auxiliary loss $\mathcal{L}_{\text{aux}}$ that emphasizes fine-grained details. The final objective is
\begin{equation}
\begin{split}
\mathcal{L}_{\text{total}} = & \; \lambda_{\text{mask}}\mathcal{L}_{\text{mask}}
+ \lambda_{\text{dice}}\mathcal{L}_{\text{dice}} \\
& + \lambda_{\text{iou}}\mathcal{L}_{\text{iou}}
+ \lambda_{\text{direct}}\mathcal{L}_{\text{direct}}
+ \lambda_{\text{aux}}\mathcal{L}_{\text{aux}},
\end{split}
\label{eq:total_loss}
\end{equation}
with $\lambda_{\text{iou}}{=}1.0$, $\lambda_{\text{direct}}{=}0.5$, and $\lambda_{\text{aux}}{=}0.2$.  
This extended formulation allows the model to not only generate accurate masks but also assess their quality and refine challenging object boundaries.

\begin{table*}
\begin{tabularx}{\textwidth}{XXXX}      
\toprule
      & COCO-Vitdet & COCO-YOLOv8 & COCO-Groundingdino  \\ 
\midrule
      & mAP mAP\textsuperscript{S} mAP\textsuperscript{M} mAP\textsuperscript{L} &  mAP mAP\textsuperscript{S} mAP\textsuperscript{M} mAP\textsuperscript{L} & mAP mAP\textsuperscript{S} mAP\textsuperscript{M} mAP\textsuperscript{L}\\
\midrule
{\scriptsize EfficientViT-SAM-L0 \cite{zhang2024efficientvit}}&  42.3\hspace{1em}25.3\hspace{1em}45.7\hspace{1em}60.5 & 40.4\hspace{1em}22.5\hspace{1em}43.8\hspace{1em}59.0 &
42.9\hspace{1em}26.8\hspace{1em}46.7\hspace{1em}62.8\\
\midrule
{\scriptsize EfficientViT-SAM-L1 \cite{zhang2024efficientvit}}&  43.6\hspace{1em}26.3\hspace{1em}47.1\hspace{1em}61.3 & 41.3\hspace{1em}23.1\hspace{1em}44.9\hspace{1em}59.7 &
43.9\hspace{1em}27.6\hspace{1em}48.2\hspace{1em}63.4\\
\midrule
{\scriptsize EfficientViT-SAM-L2 \cite{zhang2024efficientvit}}&  44.0\hspace{1em}26.9\hspace{1em}48.1\hspace{1em}61.3 & \textbf{41.9}\hspace{1em}23.6\hspace{1em}\textbf{45.9}\hspace{1em}\textbf{59.8} &
44.5\hspace{1em}28.2\hspace{1em}\textbf{49.2}\hspace{1em}\textbf{63.6}\\
\midrule
{\scriptsize EfficientViT-L2-Depth (ours)}&  \textbf{44.3}\hspace{1em}\textbf{27.8}\hspace{1em}\textbf{48.2}\hspace{1em}61.3 & 41.8\hspace{1em}\textbf{24.0}\hspace{1em}45.8\hspace{1em}59.7 &
\textbf{44.8}\hspace{1em}\textbf{29.0}\hspace{1em}\textbf{49.2}\hspace{1em}63.4\\

\bottomrule

\end{tabularx}
\caption{Zero-shot instance segmentation results using bounding boxes from ViTDet, YOLOv8, and Grounding DINO.}
\label{tab:Dino}
\end{table*}

\section{Experiments}
\label{sec:Experiments}

\subsection{Experimental Settings}
We build on EfficientViT-SAM-L2, which offers the best trade-off between accuracy and computation cost, and extend it with our depth-aware framework. Training is conducted in two stages. In the first stage, the depth encoder is randomly initialized and trained independently for 2 epochs on the SA-1B example dataset (11.2k images). In the second stage, the entire model is trained end-to-end using the full objective in Eq.~\ref{eq:total_loss}. Following the original EfficientViT-SAM setup, we sample 4 masks per image. 

Overall, the model is trained for 4 epochs on 11.2k images with a batch size of 4 using the AdamW optimizer, with momentum parameters $\beta_1=0.9$ and $\beta_2=0.999$. Loss weights are set to $\lambda_{\text{mask}}{=}20$, $\lambda_{\text{dice}}{=}1$, $\lambda_{\text{iou}}{=}1.0$, $\lambda_{\text{direct}}{=}0.5$, and $\lambda_{\text{aux}}{=}0.2$. Training is performed on two NVIDIA A6000 GPUs and completes in under 5 hours.

\subsection{Runtime Efficiency}
On a single NVIDIA A6000 GPU (fp16, no TensorRT), EfficientViT-SAM-L2 runs at 62.8 images/s on $1024{\times}1024$ inputs, while our depth-aware extension achieves 31.9 images/s with roughly doubled parameters (61.3M $\rightarrow$ 118.7M) and MACs (69G $\rightarrow$ 137G), yet remains far smaller than SAM-ViT-H ($>$600M, $\sim$3000G) and maintains practical efficiency with improved accuracy.

\subsection{Qualitative Results}
Figure~\ref{fig:comparison} presents qualitative comparisons. Our method produces noticeably sharper and more consistent segmentation masks, especially around fine structures, object boundaries, and small objects that RGB-only models struggle with. This improvement arises because the depth map provides complementary geometric information, enabling the model to better capture boundaries and separate closely adjacent objects. These visual results highlight the benefit of integrating depth features into the segmentation process.

\subsection{Zero-Shot Box-Prompted Segmentation}
We evaluate in the box-prompted setting using ground-truth bounding boxes on COCO \cite{lin2014microsoft} and LVIS \cite{gupta2019lvis}, where our method consistently surpasses EfficientViT-SAM-L2 in mean IoU, with notable gains on small and medium objects (Table~\ref{tab:GT-box}). To assess real-world applicability, we further test with automatically generated boxes from ViTDet \cite{li2022exploring}, YOLOv8 \cite{wang2023uav}, and Grounding DINO \cite{liu2024grounding} for text-guided proposals, and again observe consistent improvements, particularly on small objects (Table~\ref{tab:Dino}). An ablation confirms this trend, as removing depth fusion reduces accuracy by about 1\% on small objects, highlighting the contribution of our depth-aware design.

\begin{table}
\begin{tabularx}{\linewidth}{l l l}      
\toprule
      & COCO & LVIS  \\ 
\midrule
Click times     &{\small 1 \hspace{2em}  3 \hspace{2em}  5 } & {\small 1\hspace{2em} 3\hspace{2em} 5}\\
\midrule
SAM-ViT-H  \cite{kirillov2023segment}& {\small 58.4\hspace{1em}69.6\hspace{1em}71.4} & {\small \textbf{59.2}\hspace{1em}66.0\hspace{1em}66.8}\\
\midrule
{\scriptsize EfficientViT-SAM-L2 \cite{zhang2024efficientvit}} & {\small 56.1\hspace{1em}67.2\hspace{1em}69.4} & {\small 57.8\hspace{1em}64.9\hspace{1em}66.1}\\
\midrule
{\scriptsize EfficientViT-L2-Depth (ours) } & {\small 52.6\hspace{1em}\textbf{71.6}\hspace{1em}\textbf{75.5}} & {\small 56.3\hspace{1em}\textbf{70.4}\hspace{1em}\textbf{73.8}}\\
\midrule
{\scriptsize EfficientViT-SAM-XL1 \cite{zhang2024efficientvit}}& {\small \textbf{59.8}\hspace{1em}71.3\hspace{1em}75.3} & {\small 56.6\hspace{1em}67.0\hspace{1em}71.7}\\
\bottomrule
\end{tabularx}
\caption{Zero-shot point-prompted results (mIoU) on COCO and LVIS with 1, 3, and 5 clicks.}
\label{tab:point-prompted}
\end{table}

\subsection{Zero-Shot Point-Prompted Segmentation}
We next evaluate in the point-prompted setting, where performance is measured with 1, 3, and 5 clicks per object using mean IoU as the metric. Table~\ref{tab:point-prompted} summarizes the results. Our method achieves clear and consistent gains, particularly when more user clicks are provided. With 3 and 5 clicks, our model outperforms EfficientViT-SAM-L2, the larger EfficientViT-SAM-XL1, and even the original SAM-ViT-H. For instance, on LVIS with 5 clicks, our method improves over SAM-ViT-H by \textbf{7.7\%} and surpasses EfficientViT-SAM-XL1 by \textbf{2.1\%}. These findings demonstrate that incorporating depth information allows the model to better leverage sparse prompts, yielding more accurate masks and stronger cross-dataset generalization.

%\vfill
%\pagebreak

\begin{table}[t]
\centering
\resizebox{\columnwidth}{!}{%
\begin{tabular}{lcc}
\toprule
& COCO & LVIS \\
\midrule
& {\scriptsize mIoU \quad mIoU\textsuperscript{S} \quad mIoU\textsuperscript{M} \quad mIoU\textsuperscript{L}} 
& {\scriptsize mIoU \quad mIoU\textsuperscript{S} \quad mIoU\textsuperscript{M} \quad mIoU\textsuperscript{L}} \\
\midrule
SAM-ViT-H \cite{kirillov2023segment} & 77.4 \; 72.3 \; 80.4 \; 81.8 & 77.0 \; 70.6 \; 87.5 \; 89.9 \\
{\scriptsize \mbox{EfficientViT-SAM-L2 \cite{zhang2024efficientvit}}} & 75.6 \; 69.7 \; 78.7 \; \textbf{81.7} & 75.5 \; 68.6 \; 84.9 \; \textbf{88.8} \\
{\scriptsize \mbox{EfficientViT-L2-Depth (ours)}} & \textbf{76.6} \; \textbf{72.2} \; \textbf{78.9} \; 81.2 & \textbf{76.7} \; \textbf{70.6} \; \textbf{85.1} \; 88.6 \\
{\scriptsize \mbox{EfficientViT-SAM-XL1 \cite{zhang2024efficientvit}}} & 79.9 \; 75.8 \; 82.2 \; 83.7 & 79.9 \; 74.4 \; 88.4 \; 91.6 \\
\bottomrule
\end{tabular}%
}
\caption{Zero-shot instance segmentation results using ground-truth bounding boxes, with results reported for small, medium, and large objects.}
\label{tab:GT-box}
\end{table}

\section{Conclusion}
In this work, we introduced Depth-Aware EfficientViT-SAM, a lightweight framework that integrates monocular depth with EfficientViT-SAM through a dedicated encoder. This fusion improves segmentation quality, especially on object boundaries and small objects, and in point-prompted settings surpasses larger variants such as EfficientViT-SAM-XL1 and SAM, all while training on only 11.2k images—a tiny fraction of SA-1B—demonstrating strong data efficiency.

Although the depth branch roughly doubles parameters and computation compared to EfficientViT-SAM-L2, the model remains lightweight relative to SAM-ViT-H and achieves consistent accuracy gains with limited training data. Importantly, our runtime analysis shows that training can be completed in only a few hours on commodity GPUs such as the A6000, making the approach practical for research and deployment without requiring extreme multi-A100 infrastructure. Overall, our work shows that depth-aware fusion is an effective way to enhance SAM-based frameworks, enabling practical, geometry-aware, and data-efficient segmentation.

% \vfill\pagebreak

\bibliographystyle{IEEEbib}
\bibliography{strings,refs}

@inproceedings{kirillov2023segment,
  title={Segment anything},
  author={Kirillov, Alexander and Mintun, Eric and Ravi, Nikhila and Mao, Hanzi and Rolland, Chloe and Gustafson, Laura and Xiao, Tete and Whitehead, Spencer and Berg, Alexander C and Lo, Wan-Yen and others},
  booktitle={Proceedings of the IEEE/CVF international conference on computer vision},
  pages={4015--4026},
  year={2023}
}

@article{zhang2023faster,
  title={Faster segment anything: Towards lightweight sam for mobile applications},
  author={Zhang, Chaoning and Han, Dongshen and Qiao, Yu and Kim, Jung Uk and Bae, Sung-Ho and Lee, Seungkyu and Hong, Choong Seon},
  journal={arXiv preprint arXiv:2306.14289},
  year={2023}
}

@article{zhao2023fast,
  title={Fast segment anything},
  author={Zhao, Xu and Ding, Wenchao and An, Yongqi and Du, Yinglong and Yu, Tao and Li, Min and Tang, Ming and Wang, Jinqiao},
  journal={arXiv preprint arXiv:2306.12156},
  year={2023}
}

@inproceedings{zhang2024efficientvit,
  title={Efficientvit-sam: Accelerated segment anything model without performance loss},
  author={Zhang, Zhuoyang and Cai, Han and Han, Song},
  booktitle={Proceedings of the IEEE/CVF Conference on Computer Vision and Pattern Recognition},
  pages={7859--7863},
  year={2024}
}

@article{cheng2023segment,
  title={Segment and track anything},
  author={Cheng, Yangming and Li, Liulei and Xu, Yuanyou and Li, Xiaodi and Yang, Zongxin and Wang, Wenguan and Yang, Yi},
  journal={arXiv preprint arXiv:2305.06558},
  year={2023}
}

@article{yin2023dformer,
  title={Dformer: Rethinking rgbd representation learning for semantic segmentation},
  author={Yin, Bowen and Zhang, Xuying and Li, Zhongyu and Liu, Li and Cheng, Ming-Ming and Hou, Qibin},
  journal={arXiv preprint arXiv:2309.09668},
  year={2023}
}

@inproceedings{yin2025dformerv2,
  title={Dformerv2: Geometry self-attention for rgbd semantic segmentation},
  author={Yin, Bo-Wen and Cao, Jiao-Long and Cheng, Ming-Ming and Hou, Qibin},
  booktitle={Proceedings of the Computer Vision and Pattern Recognition Conference},
  pages={19345--19355},
  year={2025}
}

@article{jia2024geminifusion,
  title={Geminifusion: Efficient pixel-wise multimodal fusion for vision transformer},
  author={Jia, Ding and Guo, Jianyuan and Han, Kai and Wu, Han and Zhang, Chao and Xu, Chang and Chen, Xinghao},
  journal={arXiv preprint arXiv:2406.01210},
  year={2024}
}

@inproceedings{yang2024depth,
  title={Depth anything: Unleashing the power of large-scale unlabeled data},
  author={Yang, Lihe and Kang, Bingyi and Huang, Zilong and Xu, Xiaogang and Feng, Jiashi and Zhao, Hengshuang},
  booktitle={Proceedings of the IEEE/CVF conference on computer vision and pattern recognition},
  pages={10371--10381},
  year={2024}
}

@ARTICLE{10904342,
  author={Li, Mingrui and Guo, Zhetao and Deng, Tianchen and Zhou, Yiming and Ren, Yuxiang and Wang, Hongyu},
  journal={IEEE Robotics and Automation Letters}, 
  title={DDN-SLAM: Real Time Dense Dynamic Neural Implicit SLAM}, 
  year={2025},
  volume={10},
  number={5},
  pages={4300-4307},
  doi={10.1109/LRA.2025.3546130}}

@INPROCEEDINGS{11127324,
  author={Li, Mingrui and Zhou, Yiming and Zhou, Hongxing and Hu, Xinggang and Roemer, Florian and Wang, Hongyu and Osman, Ahmad},
  booktitle={2025 IEEE International Conference on Robotics and Automation (ICRA)}, 
  title={Dy3DGS-SLAM: Monocular 3D Gaussian Splatting SLAM for Dynamic Environments}, 
  year={2025},
  volume={},
  number={},
  pages={14572-14578},
  doi={10.1109/ICRA55743.2025.11127324}}

@INPROCEEDINGS{10704527,
  author={Zhou, Yiming and Zeng, Zixuan and Chen, Andi and Zhou, Xiaofan and Ni, Haowei and Zhang, Shiyao and Li, Panfeng and Liu, Liangxi and Zheng, Mengyao and Chen, Xupeng},
  booktitle={2024 6th International Conference on Data-driven Optimization of Complex Systems (DOCS)}, 
  title={Evaluating Modern Approaches in 3D Scene Reconstruction: NeRF vs Gaussian-Based Methods}, 
  year={2024},
  volume={},
  number={},
  pages={926-931},
  doi={10.1109/DOCS63458.2024.10704527}}

@inproceedings{wang2025vggt,
  title={Vggt: Visual geometry grounded transformer},
  author={Wang, Jianyuan and Chen, Minghao and Karaev, Nikita and Vedaldi, Andrea and Rupprecht, Christian and Novotny, David},
  booktitle={Proceedings of the Computer Vision and Pattern Recognition Conference},
  pages={5294--5306},
  year={2025}
}

@inproceedings{liu2023efficientvit,
  title={Efficientvit: Memory efficient vision transformer with cascaded group attention},
  author={Liu, Xinyu and Peng, Houwen and Zheng, Ningxin and Yang, Yuqing and Hu, Han and Yuan, Yixuan},
  booktitle={Proceedings of the IEEE/CVF conference on computer vision and pattern recognition},
  pages={14420--14430},
  year={2023}
}

@inproceedings{lin2014microsoft,
  title={Microsoft coco: Common objects in context},
  author={Lin, Tsung-Yi and Maire, Michael and Belongie, Serge and Hays, James and Perona, Pietro and Ramanan, Deva and Doll{\'a}r, Piotr and Zitnick, C Lawrence},
  booktitle={European conference on computer vision},
  pages={740--755},
  year={2014},
  organization={Springer}
}

@inproceedings{gupta2019lvis,
  title={Lvis: A dataset for large vocabulary instance segmentation},
  author={Gupta, Agrim and Dollar, Piotr and Girshick, Ross},
  booktitle={Proceedings of the IEEE/CVF conference on computer vision and pattern recognition},
  pages={5356--5364},
  year={2019}
}

@article{wang2023uav,
  title={UAV-YOLOv8: A small-object-detection model based on improved YOLOv8 for UAV aerial photography scenarios},
  author={Wang, Gang and Chen, Yanfei and An, Pei and Hong, Hanyu and Hu, Jinghu and Huang, Tiange},
  journal={Sensors},
  volume={23},
  number={16},
  pages={7190},
  year={2023},
  publisher={MDPI}
}

@inproceedings{liu2024grounding,
  title={Grounding dino: Marrying dino with grounded pre-training for open-set object detection},
  author={Liu, Shilong and Zeng, Zhaoyang and Ren, Tianhe and Li, Feng and Zhang, Hao and Yang, Jie and Jiang, Qing and Li, Chunyuan and Yang, Jianwei and Su, Hang and others},
  booktitle={European conference on computer vision},
  pages={38--55},
  year={2024},
  organization={Springer}
}

@inproceedings{li2022exploring,
  title={Exploring plain vision transformer backbones for object detection},
  author={Li, Yanghao and Mao, Hanzi and Girshick, Ross and He, Kaiming},
  booktitle={European conference on computer vision},
  pages={280--296},
  year={2022},
  organization={Springer}
}

\end{document}